# On the Design of Human-Robot Collaboration Gestures

Anas Shrinah[1], Masoud S. Bahraini[2], Fahad Khan[3], Seemal Asif[3], Niels Lohse[2] and Kerstin Eder[1]

[1]School of Computer Science, University of Bristol, Bristol, UK

[2]Intelligent Automation Centre, Wolfson School of Mechanical, Electrical & Manufacturing Engineering, Loughborough University, Loughborough, UK

[3]Centre for Robotics and Assembly, Cranfield University, Cranfield, UK

**ABSTRACT**

Effective communication between humans and collaborative robots is essential for seamless Human-Robot Collaboration (HRC). In noisy industrial settings, nonverbal communication, such as gestures, plays a key role in conveying commands and information to robots efficiently. While existing literature has thoroughly examined gesture recognition and robots' responses to these gestures, there is a notable gap in exploring the design of these gestures. The criteria for creating efficient HRC gestures are scattered across numerous studies. This paper surveys the design principles of HRC gestures, as contained in the literature, aiming to consolidate a set of criteria for HRC gesture design. It also examines the methods used for designing and evaluating HRC gestures to highlight research gaps and present directions for future research in this area.

**Keywords:** Human-robot collaboration, Human-robot interaction, Gesture-based interaction, Gesture types, Gesture recognition, Gesture design, Gesture validation, robotic co-worker.

## INTRODUCTION

Human operators and robots have a complementary set of capabilities; humans perform well at tasks that require dexterity, flexibility and cognitive decision making, whereas robots are very effective in carrying out repetitive and non-ergonomic tasks (Mukherjee et al., 2022). Leveraging these special strengths requires humans and robots to share the same workspace and have a high level of collaboration. This can only be achieved if humans and robots can communicate seamlessly (Papanastasiou et al., 2019).

During collaboration, operators might need to issue commands, ask for information or give feedback to robots. These messages must be sent and received by communication channels that suit the operation environment. For instance, gestures are of great advantage in noisy environments where verbal communication might be less effective (El Makrini et al., 2018). Furthermore, gestures could augment vocal commands and provide invaluable information such as pointing towards the target object of a pick up task (Nickel and Stiefelhagen, 2007). Moreover, gestures might be of paramount importance when an operator wants an object but does not have immediate access to the name of the object or does not know its name. In such cases, humans use placeholders in their speech like "thingy" to refer to the needed item (Tárnyiková, 2019). Such vague references can be made clear to robots by using pointing gestures (pointing towards the required object) while uttering the verbal command. Therefore, gestures not only improve the interaction between humans and robots but also form an essential part of efficient and effective communication.

The great importance of gestures in Human-Robot Interaction (HRI) has led researchers to explore this area from various perspectives, including designing efficient gestures (Barattini, Morand and Robertson, 2012), developing gesture recognition technologies (Mazhar et al., 2019), and creating control algorithms for robots to respond to recognised gestures (Makris et al., 2014). Numerous reviews have surveyed the latter two areas, assessing advancements in recognition technologies (Liu and Wang, 2018) and gesture-based HRC (Wang et al., 2022). However, the literature lacks a review of the research area concerned with designing HRC.

This paper presents an analysis of the design principles for gestures used in HRC systems. Furthermore, it examines various gesture design and evaluation methods, highlighting their efficacy and limitations. Finally, it identifies existing shortcomings in the current design and validation processes of HRC gestures, offering recommendations for areas of further investigation.

This paper is organised as follows. Section 2 presents a taxonomy of HRC gestures. Section 3 consolidates a set of HRC gesture design criteria. Section 4 studies gesture evaluation methods and Section 5 concludes this paper.

## HRC GESTURE TYPES

For seamless collaboration with human workers, robots should accurately recognize human gestures and respond promptly to them. Consequently, understanding different types of gestures becomes imperative. Researchers distinguish different gesture types (Karam, 2005; Mitra and Acharya, 2007). A comprehensive understanding of these classifications is necessary to effectively implement gesture-based communication in HRC systems. Each classification offers unique insights into the nature, function, and contextual relevance and interpretation of gestures, thereby aiding in the design and development of intuitive, efficient and effective HRC interfaces. In this section, we examine the various types of gestures employed in HRC and we present a taxonomy of these gesture, as depicted in Figure 1.

1. *Modality*: Gestures serve as a fundamental means of non-verbal communication and can be expressed through the hands and arms (manual) or other body parts like the head, body, face, and gaze (non-manual). Manual gestures can be classified into static and dynamic. Static gestures involve fixed final poses, exemplified by actions like pointing, which convey clear and decisive commands or indications (Li and Zhang, 2022). On the other hand, dynamic gestures involve motions, such as waving, which often convey continuous or evolving messages (Xu et al., 2015). Non-manual gestures, meanwhile, encompass a diverse range of communicative cues: head nods, body posture shifts, facial expressions, and eye movements (Calisgan et al., 2012; Mitra and Acharya, 2007; Urakami and Seaborn, 2023).

2. *Function:* Gestures play an important role in facilitating HRC by enhancing communication and interaction between humans and robots. Classifying gestures into informative, descriptive, expressive, and regulatory categories provides a structured understanding of their roles. Informative gestures serve as direct communicative signals such as "stop" or "start", ensuring clear and efficient communication between human workers and robotic co-workers. Descriptive gestures provide visual aids for conveying information regarding objects, locations, or actions pertinent to the collaborative task at hand. Expressive gestures imbue interactions with emotional depth and nuance, allowing individuals to convey feelings or attitudes through non-verbal means. Whether signalling approval with a "thumbs up" or expressing uncertainty with a shrug, expressive gestures foster empathy and rapport between human and robotic co-workers. Regulatory gestures exert influence over robot movement or behaviour, dictating operational parameters or task priorities within the collaborative framework (Mazhar et al., 2019; Sheikholeslami, Moon and Croft, 2017).

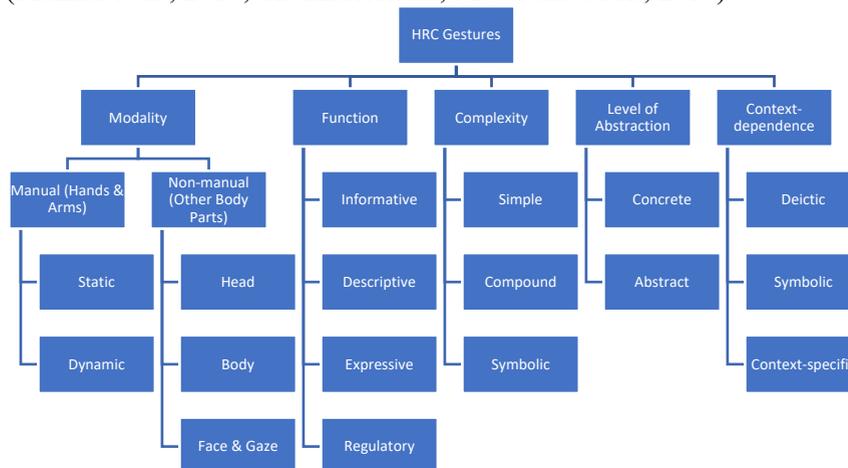

*Figure 1 Taxonomy of HRC gestures*

3. *Complexity:* At one end of the spectrum of HRI, there are simple gestures, characterized by their straightforward and easily recognizable nature. These gestures, such as pointing to indicate direction or an object of interest, serve as fundamental building blocks of non-verbal communication, offering clear and unambiguous cues to recipients (Urakami and Seaborn, 2023). Moving along the continuum, compound gestures emerge, combining multiple body parts or movements to convey more nuanced messages. For instance, a gesture that involves both pointing and nodding simultaneously not only directs attention but also affirms or acknowledges agreement, adding layers of meaning and intentionality to the interaction (Mukherjee et al., 2022). Symbolic gestures occupy the highest echelon of complexity, drawing upon cultural or contextual associations to imbue them with significance beyond their physical manifestation (Sheikholeslami, Moon and Croft, 2017). Examples of symbolic gestures include the universally recognized "peace sign", which conveys notions of peace, harmony, and solidarity extending beyond linguistic and cultural barriers.

4. *Level of Abstraction:* In addition to complexity, gestures can be classified based on their semantics which range from concrete representations of physical actions or objects to abstract symbols conveying nuanced concepts. Concrete gestures directly correspond to actions or objects, facilitating clear communication. For example, mimicking grasping or pouring actions provides straightforward cues for task coordination. Abstract gestures transcend immediate physicality, symbolizing complex ideas through metaphorical associations. For instance, extending open palms upward symbolizes openness or vulnerability, enriching interpersonal communication with profound symbolic meaning (Mazhar et al., 2019; Sheikholeslami, Moon and Croft, 2017).

5. *Context-dependence:* Gestures vary in their reliance on context, impacting how they are interpreted and their significance in interactions. Thus, gestures are classified into deictic, symbolic and context-specific. Deictic gestures are closely linked to specific objects or locations, providing clear references within the immediate environment, as seen in pointing. Symbolic gestures, like the "thumbs up," have universal meanings across cultures, transcending linguistic barriers. Context-specific gestures, such as factory hand signals, are tailored to specific tasks or environments, carrying specialized significance. Understanding the context-dependence of gestures enables effective communication and interaction across diverse settings (Karam, 2005; Mitra and Acharya, 2007; Sheikholeslami, Moon and Croft, 2017).

The taxonomy of gesture depicted in Figure 1 serves as a foundational framework for designing intuitive and efficient HRC interfaces. By exploring the nuances of gesture classification, researchers and designers can ensure seamless communication and interaction between human workers and robotic co-workers.

## GESTURE DESIGN CRITERIA

The design of HRC gestures is crucial for enhancing communication and interaction between humans and robots. This section synthesises insights from the literature to establish key criteria for designing effective HRC gestures.

### Key Criteria for Designing HRC Gestures

1. *Detection Rate:* Gestures should be designed to be easily recognised by robots. For each detection method and application, some gestures might be more robust than others, i.e. have a high recognition rate. For instance, the thumbs up gesture has a higher recognition rate than open palm (Gupta et al., 2016).

2. *Intuitiveness:* Gestures should be intuitive, allowing users to understand and perform them with minimal training. For example, pointing directly builds upon the human tendency to indicate objects of interest with our hands. Intuitive gestures facilitate their adoption hence contributing to more efficient human-robot collaboration (Nielsen et al., 2004).

3. *Ergonomics:* Gestures should not require awkward postures or excessive force, which could lead to physical strain or injury over time. For instance, Tang and Webb (2018) analysed the movement of operators' arms and hands while performing control gestures and reported that ergonomic gestures should not cause operators to move their arms over of 20 degrees from the rest position.

4. *Distinguishability:* Gestures must be easily distinguishable from one another as substantial similarities may cause confusion (Papanagiotou, Senteri and Manitsaris, 2021). A control gesture should be distinct from other control gestures, task-related actions, and spontaneous human motions (Barattini, Morand and Robertson, 2012). This gesture design criterion is very important, because human spontaneous actions frequently occur and could easily be mistaken for e.g. control gestures in manufacturing scenarios where humans are expected to collaborate with robots (Pohlt et al., 2017).

5. *Social Acceptability:* Social acceptability pertains to the design of gestures that are appropriate and non-offensive in a given cultural context. For instance, the gesture of holding out the hand with the palm up might be understood as a way of handing over the turn to a robotic co-worker. However, in South America, this gesture could be perceived as rude and offensive, implying that someone is stupid or incompetent (Lefevre, 2011).

6. *Simplicity:* Simple gestures are preferred over complex ones in HRC due to their straightforward nature and ease of recognition. The complexity of gestures can hamper their consistent execution and reduce recognition accuracy (Wachs et al., 2011). For instance, static one-handed gestures are often preferable to complex multi-step motions or full-body gestures (Shukla, Erkent and Piater, 2018).

7. *Contextual Appropriateness:* Gestures should be designed with consideration for contextual relevance and interpretation to ensure they are appropriate and carry the intended meaning. For instance, gestures designed for underwater collaboration should consider that while the "thumbs up" gesture usually signifies a positive confirmation, in diving scenarios, this gesture signals the intention to end the dive and rise to the surface (Riccardi and Desai, 2022).

8. *Mixed teams gestures:* In the case of mixed teams, gestures designed to communicate with robot and human co-workers must be the same, i.e. no matter whether the worker uses gestures when working with another human or a robotic co-worker, the gestures should ideally be the same. That is because the cognitive load involved in learning two sets of gestures is too high in most cases. This can lead to confusion, mistakes, and eventually safety hazards (Bustillos et al., 2019).

The HRC gesture design requires careful consideration of a comprehensive set of criteria to ensure effective communication between humans and robots.

## DESIGNING AND VALIDATING HRC GESTURES

The design and validation of HRC gestures are critical parts of the effort to achieve seamless and efficient interaction between humans and robots. As discussed in Section 3, gestures must be carefully designed to be both natural for humans and easily interpretable by robots. Furthermore, to ensure gestures have a high detection rate, and are intuitive, ergonomic and socially acceptable, gesture validation must combine empirical experiments with user-centred analysis in a variety of anticipated environments. This section describes design and validation methodologies of HRC gestures and discusses their strengths and weaknesses.

### HRC Gesture Design Methods

Many approaches are used design HRC gestures. One method involves observing Human-Human Interactions (HHI) to extract intuitive gestures (Calisgan et al., 2012; Gleeson et al., 2013; Pohlt et al., 2017). Though defining gestures by observing HHI is a useful approach to designing intuitive and perhaps ergonomic gestures, such methods do not consider the accuracy a robot can achieve in recognising these gestures (Gupta et al., 2016). Thus, gestures produced by such methods must be refined by other design layers to ensure gestures are easy to be recognised by robots.

Another method is to adopt well-defined gestures like the gestures of the American Sign Language for issuing commands to robots (Ding and Su, 2023; Mazhar et al., 2019). The advantage of using gesture languages in designing HRC gestures is that these gestures are established and well-defined. However, sign languages are complicated and not easy to learn. This makes the gestures of sign languages less attractive options for HRC gestures, where a relatively small set of simple and intuitive gestures are needed to support seamless collaboration (Barattini, Morand and Robertson, 2012).

In addition to the previous methods, some researchers took requirement engineering approaches for designing HRC gestures. For example, systematic requirement-gathering methods, like user journals, are used to define and evaluate a set of gestures that are natural and suitable for the intended application (Prati et al., 2021). In another research, gestures' design requirements were extracted by reviewing work environment requirements and general HRI design concepts. Then, these gesture requirements are used to propose a set of gestures (Barattini, Morand and Robertson, 2012). Leveraging requirement engineering techniques to design HRC gestures helps engineers to design gestures with user-centered and methodological approaches. Requirement engineering approaches produce gestures that are customised to user needs and tailored to the intended application. Such gestures enhance user experience and, therefore, increase collaboration efficiency. However, since requirement engineering approaches use comprehensive design processes, these methods tend to be time consuming and resource intensive. Furthermore, such methods may inadvertently introduce user bias if extra attention is not given to the data gathering step.

Commonly used HRC gestures can be extracted from the literature (Terreran, Barcellona and Ghidoni, 2023). This method offers quick access to a range of gestures frequently used in the field. However, the effectiveness of these gestures in a given context may be limited as they may be only have been used for demonstrating gesture recognition algorithms and gesture-based robot collaboration systems (El Makrini et al., 2018; Tsarouchi et al., 2017).

Each of these gesture designing approaches studies the HRC gesture design from its perspective. However, gesture design is a multidisciplinary task and needs collaboration between roboticists, computer scientists and social scientists. Hence, interdisciplinary research is needed to design effective HRC gestures.

**HRC Gesture Validation**

Assessment of gestures employs different validation methods depending on the intended evaluation criterion. One of the most important properties of gestures is their intuitiveness. Gestures must be intuitive so they are easy to learn and use by operators. Gesture intuitiveness is best validated with subjective measures like questionnaires. For instance, Gleeson et al. (2013) proposed a lexicon of gestures and invited participants to watch recordings of a human operator collaborating with a robot using these gestures. The participants are then asked to provide their subjective opinion about the naturalness and easiness of the proposed gestures.

In addition to intuitiveness, gestures must be easy to detect by robots. In the literature, this property is mostly associated with the gesture recognition systems rather than with the gestures themselves (Qi et al., 2023). However, gestures can be classified based on their miss-classification rates independently from recognition systems. For example, Gupta et al. (2016) calculated the recognition rate of 25 hand automotive user-interface gestures using five vision based classifiers and showed that some gestures constantly are more accurately recognised than others. HRC gesture design must also consider and evaluate the recognition rate of proposed gestures to ensure gestures can be accurately recognised independently from the recognition technique.

Another important attribute of gestures is their ergonomics. Ergonomic gestures prevent physical strain and enhance productivity. Gesture ergonomics can be assessed using postural analysis to measure the risk of musculoskeletal injury associated with the usage of gestures (Tang and Webb, 2018).

Social acceptance is investigated in HRI (Rico and Brewster, 2010; Xia et al., 2022). Rico and Brewster (2010) measured the social acceptance of a set of gestures by showing these gestures to participants and asking them whether they would feel comfortable performing these gestures in given locations and in front of specific audiences. Evaluating the social acceptance of gestures in different cultural contexts is extremely important because what might be seen as an innocent gesture in one culture could be rude in another (Lefevre, 2011).

While the HRC field recognises the significance of validating the intuitiveness and ergonomics of gestures, the exploration of gestures' recognition rates and social acceptance is less developed. Thus, advancing the research frontiers in these areas is essential to complement the current validation methods of HRC gestures.

## CONCLUSION

The fundamental role of gestures in improving communication between humans and robotic co-workers motivates the need for a deep investigation of HRC gestures. To help researchers develop a more holistic understanding of the factors that influence HRC gesture design, this paper classified gestures based on their modality, functionality, complexity, level of abstraction, and usage context. Additionally, this paper discussed the importance of considering the gestures' detection rate, intuitiveness, ergonomics, social acceptance, complexity, and context while designing HRC gestures. Furthermore, this study examined gesture design methods and highlighted their strengths and weaknesses. For instance, while observing HHIs produces intuitive and ergonomic gestures, such approaches often overlook the critical aspect of recognition accuracy of these gestures by robots. Sign languages are well-defined but are complex and not easily adaptable for HRC's needs. Requirement engineering techniques offer user-centered designs but are time-intensive and risk introducing user bias. Lastly, surveying gesture-recognition literature provides quick access to common gestures, but these gestures often lack in-depth validation for their suitability for HRC applications. Evidently, an interdisciplinary approach is required to design effective HRC gestures. Moreover, by reviewing HRC gesture validation methods, this paper indicated the significance of validating gestures' intuitiveness and ergonomics and pointed to the need to advance the validation of gestures' recognition rates and social acceptance in HRC applications.

## ACKNOWLEDGMENT

This work was supported by the EPSRC under Grant EP/V050966/1.

**REFERENCES**

Barattini, P., Morand, C. and Robertson, N.M. (2012) 'A proposed gesture set for the control of industrial collaborative robots', *2012 IEEE RO-MAN: The 21st IEEE International Symposium on Robot and Human Interactive Communication*. IEEE, pp. 132–137.

Bustillos, M.A.B., Maldonado-Macías, A.A., García-Alcaraz, J.L., Arellano, J.L.H. and Sosa, L.A. (2019) 'Considerations of the Mental Workload in Socio-Technical Systems in the Manufacturing Industry: A Literature Review', *Advanced Macroergonomics and Sociotechnical Approaches for Optimal Organizational Performance*, IGi Global, pp. 99–116.

Calisgan, E., Haddadi, A., Van der Loos, H.F.M., Alcazar, J.A. and Croft, E.A. (2012) 'Identifying nonverbal cues for automated human-robot turn-taking', *2012 IEEE RO-MAN: The 21st IEEE International Symposium on Robot and Human Interactive Communication*. IEEE, pp. 418–423.

Ding, I.-J. and Su, J.-L. (2023) 'Designs of human–robot interaction using depth sensor-based hand gesture communication for smart material-handling robot operations', *Proceedings of the Institution of Mechanical Engineers, Part B: Journal of Engineering Manufacture*, 237(3) SAGE Publications Sage UK: London, England, pp. 392–413.

Gleeson, B., MacLean, K., Haddadi, A., Croft, E. and Alcazar, J. (2013) 'Gestures for industry intuitive human-robot communication from human observation', *2013 8th ACM/IEEE International Conference on Human-Robot Interaction (HRI)*. IEEE, pp. 349–356.

Gupta, S., Molchanov, P., Yang, X., Kim, K., Tyree, S. and Kautz, J. (2016) 'Towards selecting robust hand gestures for automotive interfaces', *2016 IEEE Intelligent Vehicles Symposium (IV)*. IEEE, pp. 1350–1357.

Karam, M. (2005) '*A taxonomy of gestures in human computer interactions*'

Lefevre, R. (2011) *Rude hand gestures of the world: a guide to offending without words*. Chronicle Books.

Li, Y. and Zhang, P. (2022) 'Static hand gesture recognition based on hierarchical decision and classification of finger features', *Science Progress*, 105(1) SAGE Publications Sage UK: London, England, p. 00368504221086362.

Liu, H. and Wang, L. (2018) 'Gesture recognition for human-robot collaboration: A review', *International Journal of Industrial Ergonomics*, 68 Elsevier, pp. 355–367.

El Makrini, I., Elprama, S.A., Van den Bergh, J., Vanderborght, B., Knevels, A.-J., Jewell, C.I.C., Stals, F., De Coppel, G., Ravyse, I. and Potargent, J. (2018) 'Working with walt: How a cobot was developed and inserted on an auto assembly line', *IEEE Robotics & Automation Magazine*, 25(2) IEEE, pp. 51–58.

Makris, S., Tsarouchi, P., Surdilovic, D. and Krüger, J. (2014) 'Intuitive dual arm robot programming for assembly operations', *CIRP Annals*, 63(1) Elsevier, pp. 13–16.

Mazhar, O., Navarro, B., Ramdani, S., Passama, R. and Cherubini, A. (2019) 'A real-time human-robot interaction framework with robust background invariant hand gesture detection', *Robotics and Computer-Integrated Manufacturing*, 60 Elsevier, pp. 34–48.

Mitra, S. and Acharya, T. (2007) 'Gesture recognition: A survey', *IEEE Transactions on Systems, Man, and Cybernetics, Part C (Applications and Reviews)*, 37(3) IEEE, pp. 311–324.

Mukherjee, D., Gupta, K., Chang, L.H. and Najjaran, H. (2022) 'A survey of robot learning strategies for human-robot collaboration in industrial settings', *Robotics and Computer-Integrated Manufacturing*, 73 Elsevier, p. 102231.

Nickel, K. and Stiefelhagen, R. (2007) 'Visual recognition of pointing gestures for human–robot interaction', *Image and vision computing*, 25(12) Elsevier, pp. 1875–1884.

Nielsen, M., Störring, M., Moeslund, T.B. and Granum, E. (2004) 'A procedure for developing intuitive and ergonomic gesture interfaces for HCI', *Lecture Notes in Artificial Intelligence (Subseries of Lecture Notes in Computer Science)*, 2915 Springer Verlag, pp. 409–420. Available at: 10.1007/978-3-540-24598-8_38 (Accessed: 23 February 2024).

Papanagiotou, D., Senteri, G. and Manitsaris, S. (2021) 'Egocentric Gesture Recognition Using 3D Convolutional Neural Networks for the Spatiotemporal Adaptation of Collaborative Robots', *Frontiers in Neurorobotics*, 15 Frontiers Media SA, p. 703545. Available at: 10.3389/FNBOT.2021.703545 (Accessed: 23 February 2024).

Papanastasiou, S., Kousi, N., Karagiannis, P., Gkournelos, C., Papavasileiou, A., Dimoulas, K., Baris, K., Koukas, S., Michalos, G. and Makris, S. (2019) 'Towards seamless human robot collaboration: integrating multimodal interaction', *The International Journal of Advanced Manufacturing Technology*, 105 Springer, pp. 3881–3897.

Pohlt, C., Hell, S., Schlegl, T. and Wachsmuth, S. (2017) 'Impact of spontaneous human inputs during gesture based interaction on a real-world manufacturing scenario', *Proceedings of the 5th international conference on human agent interaction.*, pp. 347–351.

Prati, E., Villani, V., Grandi, F., Peruzzini, M. and Sabattini, L. (2021) 'Use of Interaction Design Methodologies for Human–Robot Collaboration in Industrial Scenarios', *IEEE Transactions on Automation Science and Engineering*, 19(4) IEEE, pp. 3126–3138.

Qi, J., Ma, L., Cui, Z. and Yu, Y. (2023) 'Computer vision-based hand gesture recognition for human-robot interaction: a review', *Complex & Intelligent Systems*, Springer, pp. 1–26.

Riccardi, N. and Desai, R.H. (2022) 'Discourse and the brain : Capturing meaning in the wild', *The Routledge Handbook of Semiosis and the Brain*,

Routledge, pp. 174–189. Available at: 10.4324/9781003051817-14 (Accessed: 27 February 2024).

Rico, J. and Brewster, S. (2010) 'Usable gestures for mobile interfaces: evaluating social acceptability', *Proceedings of the SIGCHI Conference on Human Factors in Computing Systems.*, pp. 887–896.

Sheikholeslami, S., Moon, Aj. and Croft, E.A. (2017) 'Cooperative gestures for industry: Exploring the efficacy of robot hand configurations in expression of instructional gestures for human–robot interaction', *The International Journal of Robotics Research*, 36(5–7) SAGE Publications Sage UK: London, England, pp. 699–720.

Shukla, D., Erkent, O. and Piater, J. (2018) 'Learning semantics of gestural instructions for human-robot collaboration', *Frontiers in Neurorobotics*, 12(MAR) Frontiers Media S.A. Available at: 10.3389/FNBOT.2018.00007 (Accessed: 23 February 2024).

Tang, G. and Webb, P. (2018) 'The design and evaluation of an ergonomic contactless gesture control system for industrial robots', *Journal of Robotics*, 2018 Hindawi

Tárnyiková, J. (2019) '*English placeholders as manifestations of vague language: Their role in social interaction*'

Terreran, M., Barcellona, L. and Ghidoni, S. (2023) 'A general skeleton-based action and gesture recognition framework for human–robot collaboration', *Robotics and Autonomous Systems*, 170 Elsevier, p. 104523.

Tsarouchi, P., Matthaiakis, A.-S., Makris, S. and Chryssolouris, G. (2017) 'On a human-robot collaboration in an assembly cell', *International Journal of Computer Integrated Manufacturing*, 30(6) Taylor & Francis, pp. 580–589.

Urakami, J. and Seaborn, K. (2023) 'Nonverbal Cues in Human–Robot Interaction: A Communication Studies Perspective', *ACM Transactions on Human-Robot Interaction*, 12(2) ACM New York, NY, pp. 1–21.

Wachs, J.P., Kölsch, M., Stern, H. and Edan, Y. (2011) 'Vision-based hand-gesture applications', *Communications of the ACM*, 54(2), pp. 60–71. Available at: 10.1145/1897816.1897838 (Accessed: 23 February 2024).

Wang, X., Shen, H., Yu, H., Guo, J. and Wei, X. (2022) 'Hand and Arm Gesture-based Human-Robot Interaction: A Review', *Proceedings of the 6th International Conference on Algorithms, Computing and Systems.*, pp. 1–7.

Xia, H., Glueck, M., Annett, M., Wang, M. and Wigdor, D. (2022) 'Iteratively designing gesture vocabularies: A survey and analysis of best practices in the HCI literature', *ACM Transactions on Computer-Human Interaction (TOCHI)*, 29(4) ACM New York, NY, pp. 1–54.

Xu, D., Wu, X., Chen, Y.-L. and Xu, Y. (2015) 'Online dynamic gesture recognition for human robot interaction', *Journal of Intelligent & Robotic Systems*, 77(3–4) Springer, pp. 583–596.